\documentclass[conference]{IEEEtran}
\IEEEoverridecommandlockouts
\usepackage{cite}
\usepackage{amsmath,amssymb,amsfonts}
\usepackage{algorithm}
\usepackage{algpseudocode}
\usepackage{graphicx}
\usepackage{textcomp}
\usepackage{xcolor}
\usepackage{booktabs}
\def\BibTeX{{\rm B\kern-.05em{\sc i\kern-.025em b}\kern-.08em
    T\kern-.1667em\lower.7ex\hbox{E}\kern-.125emX}}
\begin{document}

\title{Diverse Skill Discovery for Quadruped Robots via Unsupervised Learning}

\author{
\IEEEauthorblockN{1\textsuperscript{st} Ruopeng Cui}
\IEEEauthorblockA{\textit{College of Intelligent Robotics and Advanced Manufacturing} \\
\textit{Fudan University}\\
Shanghai, China \\
23210860100@m.fudan.edu.cn}
\and
\IEEEauthorblockN{2\textsuperscript{nd} Yifei Bi}
\IEEEauthorblockA{\textit{College of Foreign Language} \\
\textit{The University of Shanghai for Science and Technology}\\
Shanghai, China \\
yifeibi@usst.edu.cn}
\and
\IEEEauthorblockN{3\textsuperscript{rd} Haojie Luo}
\IEEEauthorblockA{\textit{College of Intelligent Robotics and Advanced Manufacturing} \\
\textit{Fudan University}\\
Shanghai, China \\
hjluo23@m.fudan.edu.cn}
\and
\IEEEauthorblockN{4\textsuperscript{th} Wei Li*}
\IEEEauthorblockA{\textit{College of Intelligent Robotics and Advanced Manufacturing} \\
\textit{Fudan University}\\
Shanghai, China \\
fudan\_liwei@fudan.edu.cn}
}

\maketitle

\begin{abstract}
Reinforcement learning necessitates meticulous reward shaping by specialists to elicit target behaviors, while imitation learning relies on costly task-specific data. In contrast, unsupervised skill discovery can potentially reduce these burdens by learning a diverse repertoire of useful skills driven by intrinsic motivation. However, existing methods exhibit two key limitations: they typically rely on a single policy to master a versatile repertoire of behaviors without modeling the shared structure or distinctions among them, which results in low learning efficiency; moreover, they are susceptible to reward hacking, where the reward signal increases and converges rapidly while the learned skills display insufficient actual diversity. In this work, we introduce an Orthogonal Mixture-of-Experts (OMoE) architecture that prevents diverse behaviors from collapsing into overlapping representations, enabling a single policy to master a wide spectrum of locomotion skills. In addition, we design a multi-discriminator framework in which different discriminators operate on distinct observation spaces, effectively mitigating reward hacking. We evaluated our method on the 12-DOF Unitree A1 quadruped robot, demonstrating a diverse set of locomotion skills. Our experiments demonstrate that the proposed framework boosts training efficiency and yields an 18.3\% expansion in state-space coverage compared to the baseline.

\end{abstract}


\section{Introduction}
Deep reinforcement learning (DRL) has emerged as a powerful paradigm for enabling robotic systems to acquire diverse and complex skills, particularly in empowering quadruped robots to execute dynamic and agile movements and achieve robust locomotion control in challenging environments\cite{cao2025skeleton}. These behaviors are typically learned through the optimization of task-specific reward functions. However, the design of such functions, commonly referred to as reward engineering, is an intricate and labor-intensive process that requires meticulous calibration and balancing of various incentive and regularization components. In the context of legged robots, these functions often balance several objectives, including task performance, robustness, and energy economy. The numerous parameters render the reward-shaping process computationally intensive and time-consuming, posing a significant bottleneck for extending reinforcement learning to a wider range of applications. An alternative approach involves imitation learning (IL) methods that rely on motion capture or animatronic data, defining the reward function to encourage precise alignment between the robot's posture and the collected demonstrations. However, such approaches incur high costs, necessitating large-scale data acquisition tailored to each individual behavior. Additionally, IL must address kinematic discrepancies between the imitation data and the real robot\cite{cao2025skeleton}, and some methods based on generative adversarial imitation learning further complicate the training process by introducing multiple parameters that require fine-tuning\cite{mo2025speed, zhao2025learning}.

Unsupervised skill discovery presents a potential paradigm that eliminates the need for task-specific reward functions or additional imitation data by leveraging intrinsic motivation to autonomously explore diverse behaviors\cite{17Eysenbach2018diayn}. Current methods typically achieve this through the maximization of the mutual information between latent variables and states, where the latent variables represent skills and condition the policy\cite{17Eysenbach2018diayn,18Jiang2022recurrent}. The policy's reward signal is derived from a learned discriminator, which performs a variational approximation of the mutual information. However, existing methods typically employ either a single discriminator to distinguish states or multiple discriminators that rely on the same observation space. As a result, although the reward often converges rapidly, the discovered skills fail to adequately cover the state space and exhibit limited separability, leading to reward hacking \cite{18Jiang2022recurrent,19Campos2020explorediscoverlearn}. Moreover, existing network architectures fail to account for the shared structure and distinct characteristics among motor skills, leading to slower convergence and suboptimal performance.

To overcome these limitations, we present MOD-Skill, a novel framework that
integrates a multi-discriminator architecture and an orthogonal mixture-of-experts \cite{hendawy2023multi} policy to enhance diversity, stability, and learning efficiency in unsupervised skill discovery. Specifically, we employ multiple discriminators, each operating on a distinct observation subspace to independently assess skill diversity. Concurrently, our policy adopts an orthogonal-expert architecture in which multiple expert networks extract complementary motion features. A Gram–Schmidt process \cite{leon2013gram} enforces orthogonality among expert representations, which are subsequently combined through learned routing weights and mapped to the action space.

Our main contributions can be summarized as:
\begin{itemize}
\item We propose MOD-Skill, a novel framework that integrates an Orthogonal Mixture-of-Experts architecture with disentangled discriminators to enable efficient and diverse unsupervised skill discovery on quadruped robots.
\item The experimental results show that our multi-discriminator design effectively mitigates reward hacking, leading to an 18.3\% increase in state-space coverage. And our OMoE architecture clearly accelerates the learning process.
\item We validate the trained controller on the 12-DOF quadruped robot Unitree A1, demonstrating diverse and robust locomotion behaviors.
\end{itemize}

\section{Related Work}
\subsection{Multi-task Reinforcement Learning}
Recent research highlights modular knowledge sharing as an effective means of reducing task interference in multi-task reinforcement learning. \cite{devin2017learning} proposes a compositional policy architecture that divides the policy into modules specific to the task and the robot, enabling efficient transfer across tasks and robots and achieving zero-shot generalization to unseen robot–task combinations. \cite{sun2022paco} learns a continuous subspace of policy parameters that supports flexible parameter sharing through interpolation, yielding improved training efficiency and state-of-the-art results. \cite{d2022prioritized} proposes a multi-task reinforcement learning framework that incorporates a TD-error-based intrinsic motivation mechanism to dynamically optimize task sampling schedules, thereby effectively leveraging cross-task generalization to significantly enhance sample efficiency. In comparison, \cite{hendawy2023multi} enhances expert diversity by imposing orthogonality constraints among expert representations, providing an alternative mechanism for mitigating interference through diversified expert specialization.

\subsection{Unsupervised Skill Discovery}
Inspired by intelligent creatures' ability to interact with the physical world and learn diverse skills in the absence of external guidance, unsupervised skill discovery has emerged as a method for learning reusable skills, eschewing the need for task-specific reward functions or data. These approaches leverage intrinsic motivation, typically implemented through the maximization of the mutual information between the skill and specific aspects of the induced trajectory, enabling the learning of skill-conditioned policies capable of executing diverse and distinguishable behaviors\cite{17Eysenbach2018diayn,19Campos2020explorediscoverlearn,he2024pdrl}. \cite{tian2021unsupervised} bridges the gap between consecutive behaviors by proposing an unsupervised RL method that optimizes an information-theoretic criterion to acquire both primitive and transitional skills, facilitating the smooth composition of skill sequences.
\cite{he2024pdrl} introduces a dual representation strategy to address limited exploration and skill diversity, facilitating efficient fine-tuning for subsequent tasks.
\cite{19Campos2020explorediscoverlearn} achieves extensive state space coverage by decoupling skill acquisition from the initial state and incorporating user-specified priors.

\section{Method}
To overcome the shortcomings of existing unsupervised skill discovery algorithms, we propose a novel approach that employs multiple discriminators operating on distinct observation spaces to assess skill diversity and adopts an OMoE\cite{hendawy2023multi} architecture to extract motion features associated with different skills.

\subsection{Overview}
Within the framework of unsupervised RL, the goal is to find a family of skills indexed by the latent variable $K$, resulting in a policy $\pi_\theta(\boldsymbol{a}_t|\boldsymbol{o}^p_t,k)$ that can execute diverse state transition patterns based on latent skill $k\in\mathbb{K}$ and policy observation $\boldsymbol{o}^p$. The skill space $\mathbb{K}$ is modeled as a discrete domain containing $N_k$ distinct codes to facilitate easier human access to these skills.

\textit{1) Discriminator Module}: The discriminator module takes the motion observation $\boldsymbol{o}^m_t$ as input and produces a probability distribution over skills, estimating the likelihood that the current observation is generated by each skill. 
The motion observation $\boldsymbol{o}^m_t$ is a 33-dimensional vector, which is defined as
\begin{equation}
\boldsymbol{o}^m_t = {\left[ \boldsymbol{v}_t \quad \boldsymbol{\omega}_t \quad \boldsymbol{g}_t \quad \boldsymbol{\theta}_t \quad \dot{\boldsymbol{\theta}}_t \right]}^\top,
\end{equation}
encompassing the linear velocity $\boldsymbol{v}_t$, angular velocity $\boldsymbol{\omega}_t$, gravity direction in the body frame $\boldsymbol{g}_t$, joint positions $\boldsymbol{\theta}_t$, and joint velocities $\dot{\boldsymbol{\theta}}_t$.

\textit{2) Policy Network}:
The policy network takes the proprioceptive observation $\boldsymbol{o}^p_t$ and the skill vector $k$ as input, and outputs the action $\boldsymbol{a}_t$. The proprioceptive observation $\boldsymbol{o}^p_t$ is a 45-dimensional vector, which is defined as
\begin{equation}
\boldsymbol{o}^p_t = {\left[ \boldsymbol{o}^m_t \quad \boldsymbol{a}_{t-1} \right]}^\top,
\end{equation}
encompassing the motion observation $\boldsymbol{o}^m_t$ and the previous action $\boldsymbol{a}_{t-1}$.

\textit{3) Action Space}:
The policy action $a_t$ is a 12-dimensional vector representing offsets to the nominal joint configuration. These offsets are added to a fixed reference pose to obtain the target joint positions, which are then tracked by a joint-level PD controller with fixed gains ($K_p = 20$, $K_d = 0.5$).

\textit{4) Reward Function}:
The reward function $r$ includes the skill discovery reward $r^s$ and the regularization term $r^R$,which is defined as 
\begin{equation}
    r = \omega^S r^S + \omega^R r^R, \label{reward}
\end{equation}
where $\omega$ denotes the weighting coefficients of each reward component. In our settings, $\omega^S=\omega^R=1$. Table~\ref{regularization terms} outlines the specific formulations of the regularization terms.

\begin{table}[b]
    \caption{Regularization Terms}
    \label{regularization terms}
    \centering
    \begin{tabular}{ccccc} 
        \toprule 
        reward & equation & weight\\
        \hline
        linear velocity (z) & $\boldsymbol{v}_z^2$ & -2.0\\
        angular velocity (xy) & $\boldsymbol{\omega}_x^2 + \boldsymbol{\omega}_y^2$ & -0.05 \\
        torque & $\boldsymbol{\tau}^2$ & $-1 \times 10^{-5}$ \\
        joint acceleration & $\ddot{\boldsymbol{\theta}}^2$ &$-2.5\times 10^{-7}$ \\
        action rate & $\left( \boldsymbol{a}_t - \boldsymbol{a}_{t-1} \right)^2$ &-0.01 \\
        collision & $n_{collision}$ &-1.0 \\
        \bottomrule 
    \end{tabular}
\end{table}

\begin{figure}[t]
    \centering
    \includegraphics[width=1\linewidth]{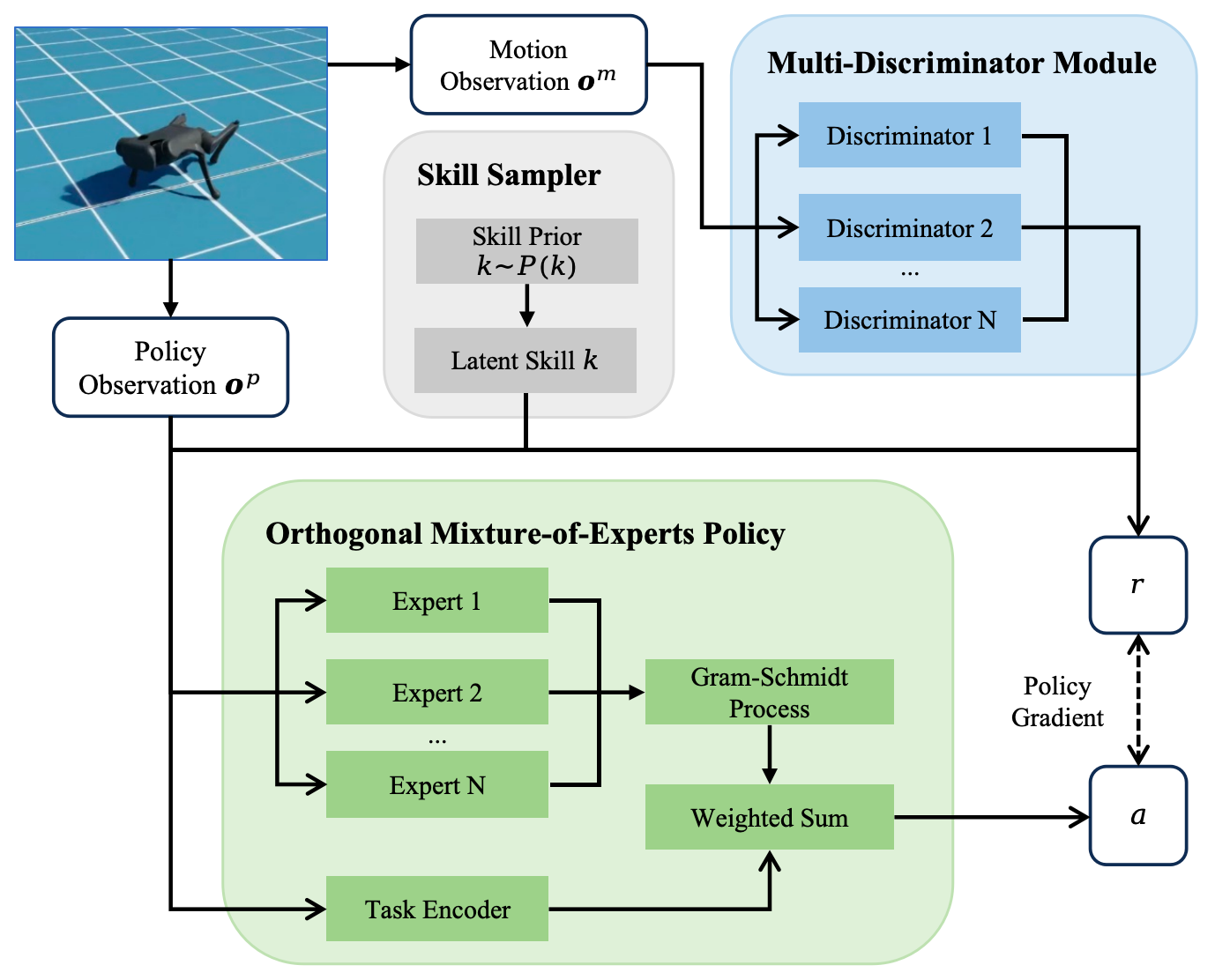}
    \caption{Overview of the proposed MOD-Skill framework. The policy is optimized through reinforcement learning, and the discriminators are updated through supervised learning, with both cooperating to facilitate skill discovery.}
    \label{fig:placeholder}

\end{figure}

\begin{algorithm}[t]
\caption{MOD-Skill}
\label{alg2}
\begin{algorithmic}[1]
\item 
    \textbf{Input:} Number of skills $N_k$, number of experts $N_e$, number of discriminators $N_d$
\item 
    \textbf{Output:} Policy $\pi_{\rm OMoE}$
\item
Initialize expert networks $E_i$, task encoder $T$, discriminator networks $q_i$ and replay buffer $M$
\For{training iteration = 1, 2, \ldots}
    \State Sample skill vector $k \sim p_k$
    \For{time step = 0, 1, 2, \ldots}
        \State Compute action $\boldsymbol{a}_t = \pi_{\rm OMoE}\left(\boldsymbol{o}^p_t,k\right)$
        \State Obtain next state $\boldsymbol{s}_{t+1} \sim p(\boldsymbol{s}_{t+1} \mid \boldsymbol{s}_t, \boldsymbol{a}_t)$
        \State Extract observation $\boldsymbol{o}_t^m$ and $\boldsymbol{o}_t^p$ from $\boldsymbol{s}_t$
        \State Compute reward $r_t$ (Eq.~\ref{reward})
        \State Store sample $\left(\boldsymbol{o}^m_t,\boldsymbol{o}^p_t, k\right)$ in the replay buffer $M$
    \EndFor
    \For{update step = 1, 2, \ldots}
        \State Sample minibatch $m \sim M$
        \State Update discriminators $q_i$ with cross-entropy loss
        \State Update value function $V$ and policy $\pi_{\rm OMoE}$ with PPO algorithm
    \EndFor
\EndFor
\end{algorithmic}
\end{algorithm}

\subsection{Multi-Discriminator Module}
A common thread across many unsupervised skill discovery methods is the design of objectives that maximize the mutual information between the latent skill $K$ and observations $O^d$ \cite{17Eysenbach2018diayn,18Jiang2022recurrent,19Campos2020explorediscoverlearn}:
\begin{equation}
\begin{aligned}
F(\theta) &= I(K,O^d)= H(K) - H(K|O^d) \\
          &= \mathbb{E}_{(k,\boldsymbol{o}^d) \sim p(k,\boldsymbol{o}^d)} \left[ \log p(k|\boldsymbol{o}^d) - \log p(k) \right]
\end{aligned}
\end{equation}
To maximize the skill entropy $H(K)$, the prior distribution $p(k)$ is fixed to be uniform. Since calculating the conditional distribution $P(K|O^d)$ is intractable, a learned parametric discriminator module $q(k|\boldsymbol{o}^d)$ is applied to approximate this posterior, which is trained to infer the underlying skill from the observations. By approximating $p$ using $q$, we derive a variational lower bound $\tilde{F}(\theta)$ with respect to $F(\theta)$:
\begin{equation}
\begin{aligned}
F(\theta) &\geq \tilde{F}(\theta)\\
 &= \mathbb{E}_{(k,\boldsymbol{o}^d) \sim p(k,\boldsymbol{o}^d)} \left[ \log q(k|\boldsymbol{o}^d) - \log p(k) \right]
\end{aligned}
\end{equation}

A multi-discriminator architecture is designed, in which each discriminator is assigned to a distinct observation subspace and computes intrinsic rewards independently, ensuring that all relevant state dimensions are properly diversified. We maximize the lower bound $\tilde{F}(\theta)$ with the following reward:
\begin{equation}
r^S_t = \frac{1}{N_d}\sum\limits_{i=1}^{N_d} \log q_i(k|\boldsymbol{o}^{d_i}_{t})-\log p(k), \label{eq:reward}
\end{equation}
where $N_d$ denotes the number of discriminators, $\boldsymbol{o}^{d_i}$ denotes the observation subspace assigned to the $i^{\rm th}$ discriminator, $\boldsymbol{o}^d = \bigcup_{i=1}^{N_d} \boldsymbol{o}^{i},\boldsymbol{o}^{d_i} \cap \boldsymbol{o}^{d_j} = \varnothing,\forall i \neq j$. To tighten the lower bound, each discriminator is updated using a cross-entropy loss so as to approximate the true posterior $p(K|O^d)$.

We employ three discriminators whose observations are
$[\boldsymbol{v}\quad  \boldsymbol{\omega}]^\top$, $[\boldsymbol{g}]^\top$, and $[\boldsymbol{\theta}\quad\dot{\boldsymbol{\theta}}]^\top$, respectively.
Each discriminator outputs a probability distribution over skills conditioned on its observation, which reduces state aliasing. 
By decoupling heterogeneous observation channels, the proposed design encourages the agent to explore a wider and richer repertoire of skills.

\subsection{Orthogonal Mixture-of-Experts Policy}
The OMoE \cite{hendawy2023multi} takes the concatenation of the proprioceptive observation $\boldsymbol{o}_t^p$ and the skill vector $k$ as input. A set of $N_e$ experts $\{E_i\}_{i=1}^{N_e}$ maps this input to feature vectors:
\begin{equation}
\boldsymbol{u}_i = E_i(\boldsymbol{o}_t^p, k)
\end{equation}

To ensure that the learned representations maintain sufficient diversity, we apply Gram–Schmidt process \cite{leon2013gram} to the generated features, constraining the outputs of the expert networks to be orthogonal:
\begin{equation}
\boldsymbol{v}_i = \boldsymbol{u}_i -\sum\limits_{j=1}^{i-1}\frac{\left<\boldsymbol{v}_j, \boldsymbol{u}_i\right>}{\left<\boldsymbol{v}_j, \boldsymbol{v}_i\right>}\boldsymbol{v}_j
\end{equation}

The task encoder $T$ computes the weighting coefficients for each expert feature:
\begin{equation}
\boldsymbol{\alpha} = T(o_t^p, k)\
\end{equation}

Subsequently, the orthogonalized expert features are aggregated through a weighted combination, and a shared output head $f$ maps the resulting representation into the action space:
\begin{equation}
\boldsymbol{a} = f\left(\sum\limits_{i=1}^{N_e}\boldsymbol{\alpha}_i\boldsymbol{v}_i\right)
\end{equation}

Leveraging the OMoE architecture, the policy decomposes diverse motion skills into a set of orthogonal representations, which are then adaptively fused by the task encoder. This enables flexible expert composition and the generation of diverse skills, improving training stability and efficiency.

\section{Experiments} 
Our method is evaluated on Unitree A1, a quadruped platform with 12 degrees of freedom. We set $N_k=100$, $N_e=6$, $N_d=3$ and train the policy with Proximal Policy Optimization (PPO,\cite{38Schulman2017ppo}) in Isaac Sim with 4096 parallel environments. The policy operates at a frequency of 50 Hz.

\begin{table}[b]
    \caption{State Space Coverage Rate}
    \label{state_coverage_discriminator_rate}
    \centering
    \begin{tabular}{ccccc} 
        \toprule 
         & SD1 & SD2 & SD3 & MD\\
        \hline
        ratio (\%) & 42.9 & 49.6 & 22.2 & 58.7\\
        \bottomrule 
    \end{tabular}
\end{table}
\begin{figure}[t]
    \centering
    \includegraphics[width=0.8\linewidth]{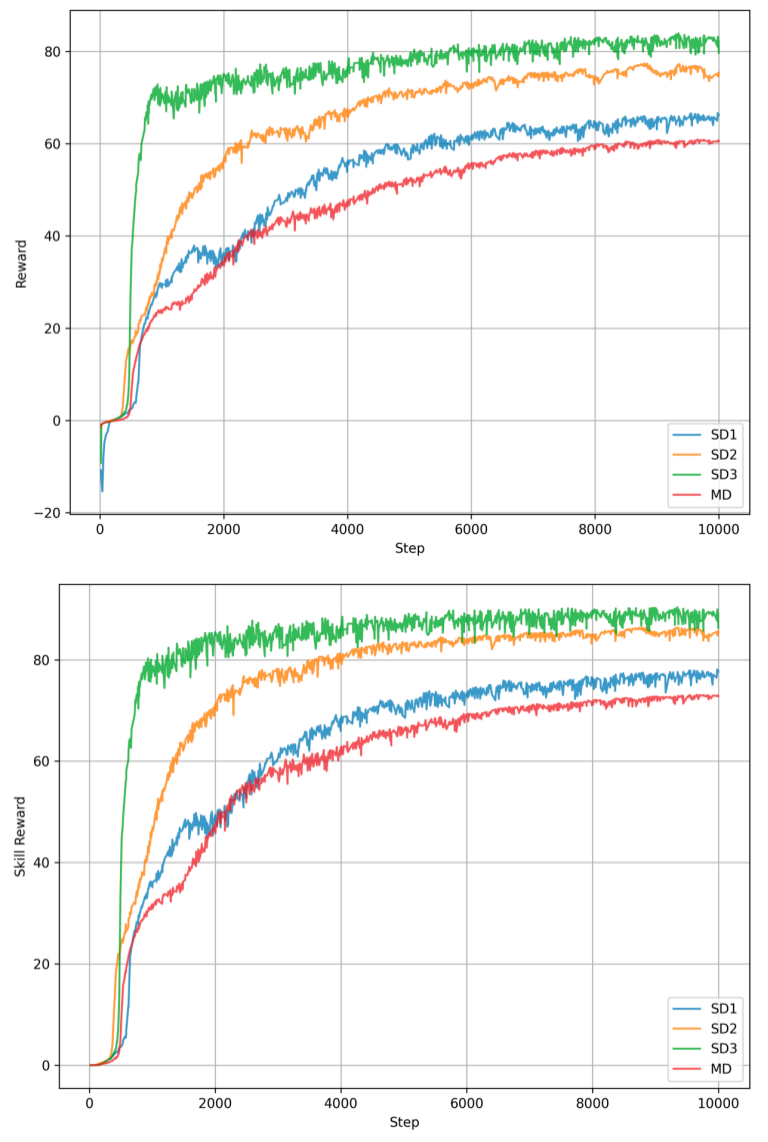}
    \caption{Reward curves for algorithm SD1, SD2, SD3, and MD. The top subplot shows the full episode reward, while the bottom subplot presents only the skill reward component.}
    \label{reward_discriminator}
\end{figure}

\begin{figure}[!h]
    \centering
    \includegraphics[width=0.8\linewidth]{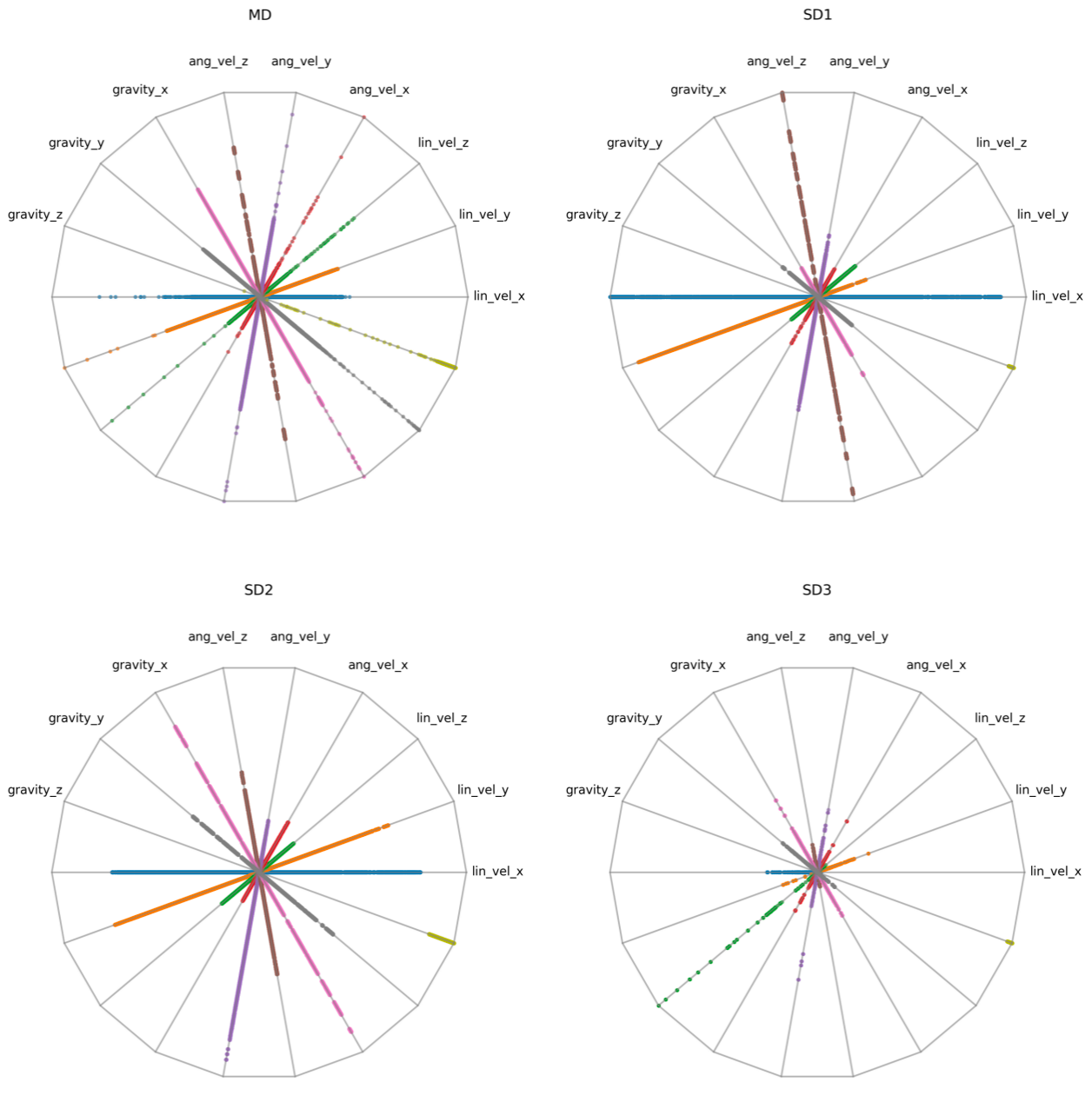}
    \caption{State Space Coverage. We roll out the skills discovered by each algorithm for 20 seconds, collect their linear velocity, angular velocity, and projected gravity observations, and normalize them by dimension.}
    \label{state_coverage_discriminator}
\end{figure}
\subsection{Ablation Study for the Discriminator Module}

The policy, value function, and discriminator module are an MLP architecture with hidden dimensions set to $[256,256,128]$. We construct three baseline methods, each following the single-discriminator (SD) architecture of DIAYN \cite{17Eysenbach2018diayn}, while differing in the observation space provided to the discriminator. Our multi-discriminator (MD) architecture uses multiple discriminators to individually assess diversity over distinct observation subspaces, thereby enabling more comprehensive diversification in the full motion-observation space. The corresponding observation configurations for each baseline and for our MD architecture are detailed as follows:
\begin{itemize}
    \item SD1: A single-discriminator model with observation space $\boldsymbol{o}^d = [\boldsymbol{v}\quad \boldsymbol{\omega}]^\top$.
    \item SD2: A single-discriminator model with extended observation space $\boldsymbol{o}^d = [\boldsymbol{v}\quad\boldsymbol{\omega}\quad\boldsymbol{g}]^\top$.
    \item SD3: A single-discriminator model with the full motion observation $\boldsymbol{o}^d = [\boldsymbol{v}\quad \boldsymbol{\omega}\quad \boldsymbol{g}\quad \boldsymbol{\theta}\quad \dot{\boldsymbol{\theta}}]^\top$.
    \item MD: A multi-discriminator model that evaluates skill diversity over multiple disjoint observation subspaces $\boldsymbol{o}^{d_1}=[\boldsymbol{v} \quad \boldsymbol{\omega}]^\top, \boldsymbol{o}^{d_2}=[\boldsymbol{g}]^\top,\boldsymbol{o}^{d_3}=[\boldsymbol{\theta} \quad \dot{\boldsymbol{\theta}}]^\top$.
\end{itemize}

\begin{figure*}[!t]
    \centering
    \includegraphics[width=1\linewidth]{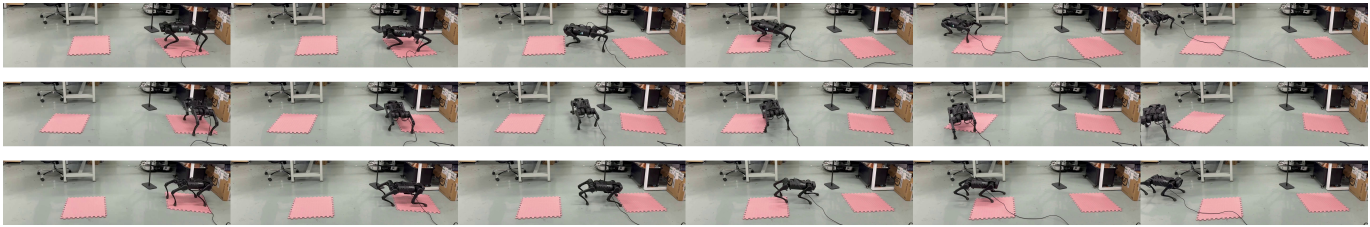}
    \caption{Real-world experiment. The learned skills are deployed on a Unitree A1 robot, demonstrating reliable and robust execution in real-world environments.}
    \label{real_world}
\end{figure*}

We train each algorithm for 10,000 iterations, and the average per-episode reward is shown in Fig.~\ref{reward_discriminator}. Subsequently, each skill is executed for 20 s in the simulation environment, during which we record its state sequence, including linear velocity, angular velocity, and projected gravity. All state sequences collected from each algorithm are normalized in every state dimension, as illustrated in Fig.~\ref{state_coverage_discriminator}.To quantify state-space coverage, each normalized state dimension is uniformly discretized into 50 bins. For each algorithm, we compute the proportion of bins that contain at least one trajectory sample, yielding the coverage ratio for that dimension. The final state-space coverage metric is obtained by averaging these proportions across all dimensions, as shown in Table~\ref{state_coverage_discriminator_rate}.

SD2 exhibits stronger diversity in the dimension of gravity compared to SD1. Although SD3 achieves the fastest reward increase, most of its skills remain stationary in different poses, demonstrating a clear reward-hacking phenomenon. In contrast, our multi-discriminator architecture achieves the largest state space range, improving the overall state-space coverage by 18.3\% compared to the best baseline algorithm, and enables the generation of locomotion skills that vary widely in terms of pose, gait, and both linear and angular velocities.

\subsection{Ablation Study for the Policy Architecture}

\begin{figure}[h]
    \centering
    \includegraphics[width=0.8\linewidth]{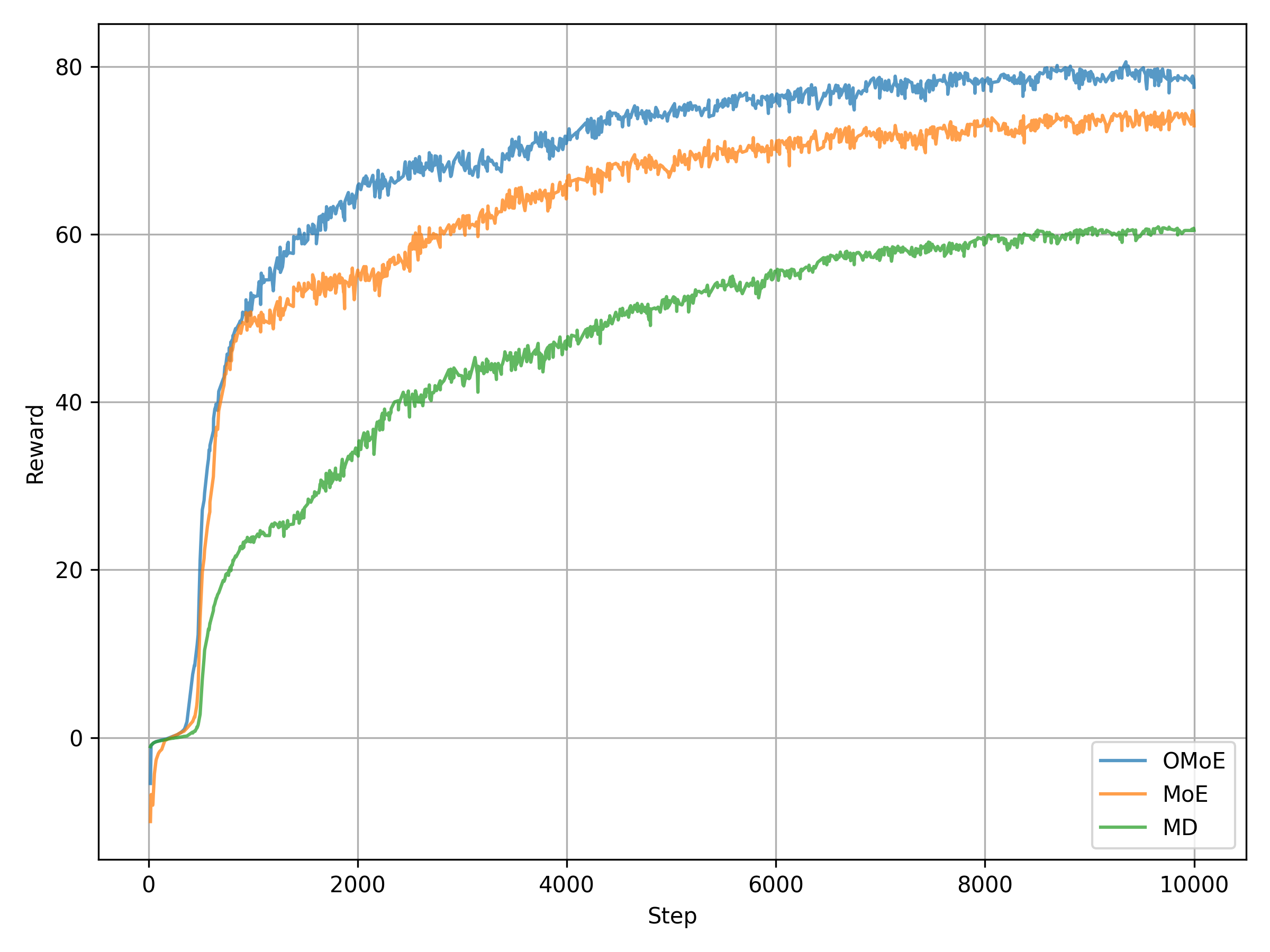}
    \caption{Reward curves for algorithm OMoE, MoE, and MLP.}
    \label{reward_omoe}
\end{figure}

We conduct an ablation study on the policy network and compare three architectures: MLP, MoE, and OMoE. For both MoE and OMoE, we employ six expert networks, where each expert is implemented as a Multi-Layer Perceptron (MLP) with hidden dimensions of $[64,64]$ and a $64$-dimensional output. The combined expert features are projected to the action space through a linear layer. To ensure a fair comparison, the configurations of MoE and OMoE are designed such that their overall parameter counts remain comparable to those of the baseline MLP. We train each algorithm for 10,000 iterations with a multi-discriminator module. As shown in Fig.~\ref{reward_omoe}, OMoE achieves a substantially faster reward increase than both MoE and MLP.

\subsection{Real-World Experiments}
Domain randomization is employed by varying training conditions to account for real-world uncertainties to bridge the sim-to-real gap. Specifically, we randomize ground friction, restitution, base and link masses, action delays, and motor torques, and introduce random velocity disturbances to simulate external force perturbations, thereby enhancing the robustness of the model. As shown in Fig.~\ref{real_world}, the learned skills can be reliably deployed in the real world.

\section{Conclusion}
In this work, we propose MOD-Skill, a novel unsupervised skill discovery framework for quadruped locomotion. By introducing a multi-discriminator design and an OMoE-based policy architecture, the framework facilitates the emergence of a broad repertoire of distinct locomotion behaviors. Experimental results show that MOD-Skill increases state-space coverage by 18.3\%, effectively mitigates reward hacking, and significantly improves training efficiency. Moreover, hardware experiments demonstrate successful sim-to-real transfer of the acquired skills to a physical quadruped robot.

In future work, we aim to advance skill discovery quality while limiting the reliance on expert knowledge, enabling algorithms to autonomously discover richer and more dynamic skill repertoires through self-guided exploration.
\bibliographystyle{IEEEtran} 
\bibliography{reference}   
\end{document}